\documentclass[sigconf, nonacm]{acmart}

\AtBeginDocument{%
  \providecommand\BibTeX{{%
    \normalfont B\kern-0.5em{\scshape i\kern-0.25em b}\kern-0.8em\TeX}}}

\setcopyright{acmcopyright}
\copyrightyear{2020}
\acmYear{2020}
\acmConference[Conference acronym 'XX]{Make sure to enter the correct
  conference title from your rights confirmation emai}{June 03--05,
  2018}{Woodstock, NY}

\usepackage{etoolbox}
\usepackage{caption}
\usepackage{subcaption}
\usepackage{multirow}
\usepackage{float}
\usepackage{tabularx}
\usepackage{adjustbox}
\usepackage{url}
\usepackage{hhline}
\usepackage{nicematrix}
\usepackage{tabularray}
\usepackage{tabularx,colortbl}
\usepackage{soul}

\newlength{\Oldarrayrulewidth}

\newcommand{\fire}{\includegraphics[height=1em]{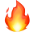} }
\newcommand{\light}{\includegraphics[height=1em]{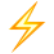} }
\newcommand{\logo}{\includegraphics[height=1em]{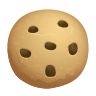}}
\newcommand{\logov}{\includegraphics[height=1em]{figures/cookie.png} }
\newcommand{\modelname}{\textsc{CooKIE }}
\newcommand{\modelnamex}{\textsc{CooKIE}}

\def\BibTeX{{\rm B\kern-.05em{\sc i\kern-.025em b}\kern-.08em
    T\kern-.1667em\lower.7ex\hbox{E}\kern-.125emX}}

\makeatletter
\patchcmd{\@classz}
  {\CT@row@color}
  {\oldCT@column@color}
  {}
  {}
\patchcmd{\@classz}
  {\CT@column@color}
  {\CT@row@color}
  {}
  {}
\patchcmd{\@classz}
  {\oldCT@column@color}
  {\CT@column@color}
  {}
  {}
\makeatother

\begin{document}

\title{Less is More: Multimodal Region Representation via Pairwise Inter-view Learning}


\author{Min Namgung}
\email{namgu007@umn.edu}
\affiliation{%
  \institution{University of Minnesota}
  \city{Minneapolis}
  \country{USA}
}

\author{Yijun Lin}
\email{lin00786@umn.edu}
\affiliation{%
  \institution{University of Minnesota}
  \city{Minneapolis}
  \country{USA}
}

\author{JangHyeon Lee}
\email{lee04588@umn.edu}
\affiliation{%
  \institution{University of Minnesota}
  \city{Minneapolis}
  \country{USA}
}

\author{Yao-Yi Chiang}
\email{yaoyi@umn.edu}
\affiliation{%
  \institution{University of Minnesota}
  \city{Minneapolis}
  \country{USA}
}

\renewcommand{\shortauthors}{Namgung et al.}

\begin{abstract}
With the increasing availability of geospatial datasets, researchers have explored region representation learning (RRL) to analyze complex region characteristics.
Recent RRL methods use contrastive learning (CL) to capture shared information between two modalities but often overlook task-relevant unique information specific to each modality. Such modality-specific details can explain region characteristics that shared information alone cannot capture.
Bringing information factorization to RRL can address this by factorizing multimodal data into shared and unique information. 
However, existing factorization approaches focus on two modalities, whereas RRL can benefit from various geospatial data.
Extending factorization beyond two modalities is non-trivial because modeling high-order relationships introduces a combinatorial number of learning objectives, increasing model complexity.
We introduce \textbf{C}r\textbf{o}ss m\textbf{o}dal \textbf{K}nowledge \textbf{I}njected \textbf{E}mbedding (CooKIE\logo), an information factorization approach for RRL that captures both shared and unique representations. CooKIE uses a pairwise inter-view learning approach that captures high-order information without modeling high-order dependency, avoiding exhaustive combinations.
We evaluate CooKIE on three regression tasks and a land use classification task in New York City and Delhi, India. Results show that CooKIE outperforms existing RRL methods and a factorized RRL model, capturing multimodal information with fewer training parameters and floating-point operations per second (FLOPs). We release the code: https://github.com/MinNamgung/CooKIE.

\end{abstract}




\maketitle
\section{Introduction}
The global urban population has grown from 52\% to 57\% over the past decade~\cite{UNpop}, increasing the demand for efficient urban planning. Socioeconomic indicators like population density and land classification support decision-making by providing insights into regional dynamics~\cite{naess2001urban, hahn2015education, dustmann2016effect}. However, these indicators often rely on costly and infrequent surveys~\cite{gebru2017using, pissourios2019survey}. As a scalable alternative, region representation learning (RRL) uses multimodal geospatial data, such as remote sensing and point-of-interest (POI) data, to predict socioeconomic indicators by transforming various geospatial data into region representations~\cite{jenkins2019unsupervised, yao2018representing, wang2017region, mvure, liu2023knowledge, li2022predicting, yan2024urbanclip, jin2024urban, xiao2024refound, zou2025deep, mai2025towards}.

Recent RRL methods use contrastive learning (CL) to model relationships between regions and between two datasets (e.g., human mobility and POI data) via intra- and inter-view learning~\cite{tkde, li2024urban}. Intra-view learning makes representations distinctive between regions within a single modality and inter-view learning aligns information across multiple datasets (i.e., modalities) for a region by focusing on shared information. 
However, inter-view CL's focus on shared information discards modality-specific (i.e., unique) information during data alignment, degrading downstream performance in various domains~\cite{factorcl}. 
Since CL-based RRL methods~\cite{liu2023knowledge, tkde, li2024urban} follow the same learning paradigm, they also risk overlooking task-relevant unique information. This limits RRL's ability to fully capture the diverse characteristics of regions, potentially decreasing downstream task performance.

FactorCL~\cite{factorcl} addresses the limitation of inter-view CL discarding unique information by introducing an information factorization approach. Specifically, FactorCL decomposes multimodal data to learn unique representations from each modality and shared information across modalities. While this approach is promising, FactorCL is designed for two modalities, limiting its applicability to RRL that benefits from multiple geospatial datasets. To overcome this, we introduce GFactorCL, a generalized form of FactorCL that handles multiple modalities for effective use in RRL.

\begin{figure*}[h]
    \centering
    \includegraphics[width=0.95\textwidth]{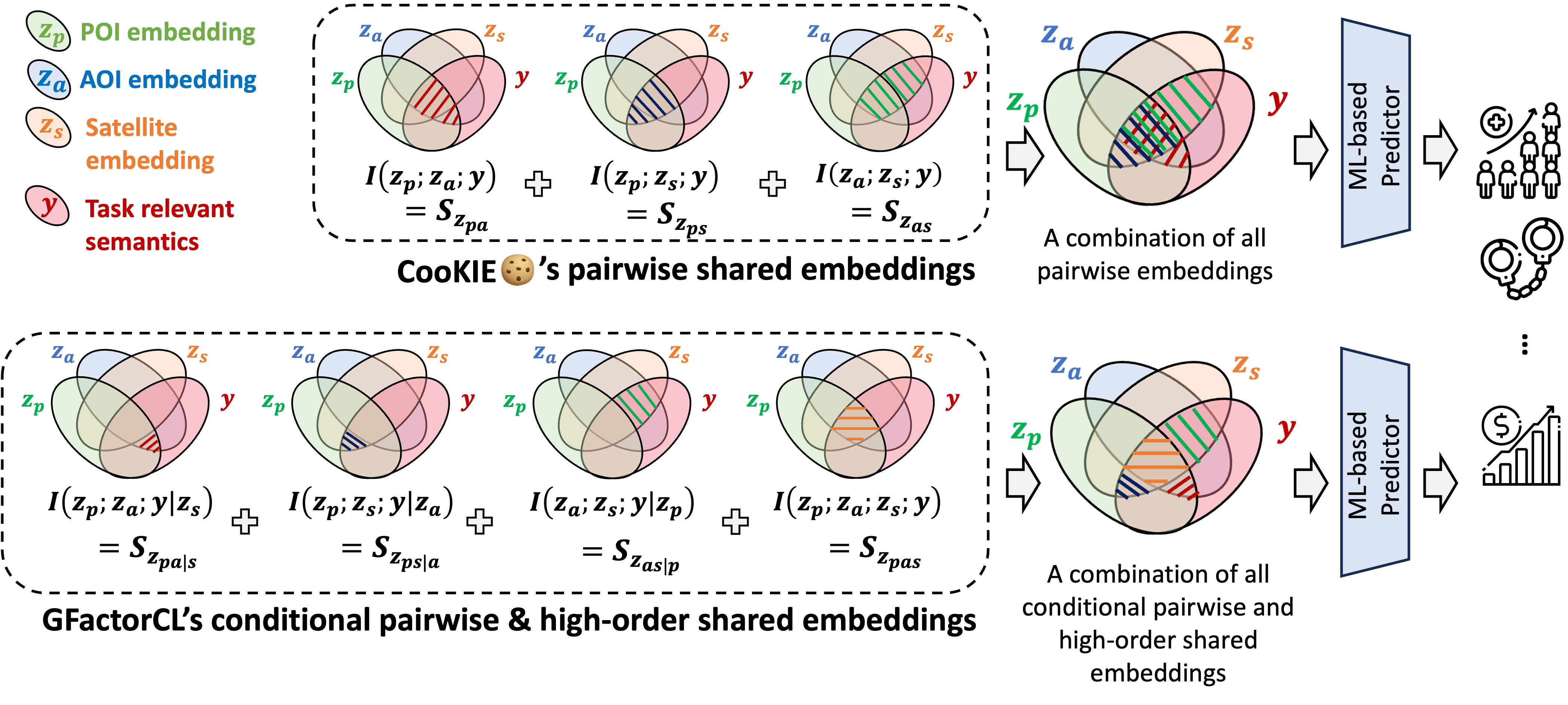}
    \vspace{0.5em}
    \caption{Comparison of shared information captured during inter-view learning between \modelnamex\logov and the direct extension of FactorCL - GFactorCL - for three modalities. \modelname captures three shared information ($S_{z_{pa}}, S_{z_{ps}}, S_{z_{as}}$) without considering conditional mutual information (CMI). GFactorCL captures three conditional pairwise information ($S_{z_{pa|s}}, S_{z_{ps|a}}, S_{z_{as|p}}$) and one high-order information ($S_{z_{pas}}$). 
    Notably, the combined pairwise embeddings in \modelname are larger than those in GFactorCL, as each shared information term captures high-order dependencies. Finally, the machine learning (ML)-based predictor utilizes all shared embeddings while handling multiple representations in predicting socioeconomic indicators.}
    \label{info}
\end{figure*}

GFactorCL can capture both shared and unique information in inter-view CL, making it a promising approach for RRL. However, the explicit factorization of high-order shared information (i.e., the shared information among $\geq 3$ modalities) in GFactorCL increases computational cost.
Approximating these high-order dependencies involves decomposing task-relevant information into a complex chain of conditional mutual information terms.
As the number of modalities grows, GFactorCL's objectives amplify, leading to an increase in training parameters and model complexity.
Instead, capturing high-order dependencies does not necessarily require conditional mutual information terms. 
According to interaction information theory~\cite{mcgill1954multivariate}, high-order dependencies among three modalities can be expressed without explicitly modeling conditional mutual information terms. Specifically, the shared information between any two modalities consists of both high-order information involving all three modalities and the conditional mutual information between the two given the third.

Therefore, we propose \modelnamex\logo,\footnote{\modelname refers to \textbf{C}r\textbf{o}ss m\textbf{o}dal \textbf{K}nowledge \textbf{I}njected \textbf{E}mbedding} a novel RRL method that applies an information factorization approach to efficiently learn task-relevant information from real-world multimodal geospatial datasets. 
\modelname addresses the computational challenge of high-order modeling by fusing modalities via a pairwise inter-view learning approach, which can learn (1) task-relevant unique information within each modality and (2) pairwise shared information between modalities without the need for exhaustive conditional dependencies. 
As shown in Figure~\ref{info}, \modelnamex's pairwise shared embeddings can capture high-order information multiple times. Even if high-order information appears multiple times, it does not increase computational cost. Moreover, the machine learning model can leverage learned representations to manage redundant information, reducing the risk of negative impact on downstream performance.

To show the effectiveness of \modelnamex, we use multiple geospatial datasets (e.g., POI data, area-of-interest (AOI) data, satellite imagery, and building footprints) to learn region representations. We evaluate these representations on regression tasks (e.g., population density prediction) and land use classification in NYC, United States, and Delhi, India. The contributions of our paper are as follows:
\begin{itemize}
    \item We bring an information factorization approach to RRL, extending FactorCL to multiple geospatial modalities via generalized FactorCL (GFactorCL). This is the first work to explicitly factorize multimodal geospatial data into shared and unique information across more than two modalities in inter-view CL, enabling effective RRL.
    \item We propose \modelnamex, a pairwise inter-view learning approach that reduces complexity in factorized multimodal RRL (i.e., GFactorCL) without explicitly modeling high-order dependencies. This approach allows \modelname to scale beyond two geospatial data while preserving both shared and unique information.
    \item We evaluate \modelname on real-world downstream tasks, including population density estimation, crime rate prediction, greenness assessment, and land use classification across heterogeneous geographical regions (NYC, USA, and Delhi, India). Results show that \modelname outperforms state-of-the-art RRL methods, reducing parameter counts by up to 56.7\ and FLOPs by up to 63\%, depending on the input modalities.
\end{itemize}

\section{\modelnamex\logo}
This section introduces \modelname for region representation learning (RRL). Given a set of regions (e.g., census tracts), \modelname aims to learn representations for each region from $m$ modalities $\mathcal{X} = \{x_1,..., x_m\}$. $Z = \{ z_1, ..., z_m\}$ represent embeddings that is derived from encoding modality $x_i$ using a backbone encoder.

\begin{figure}[h]
    \centering
\includegraphics[width=0.45\textwidth]{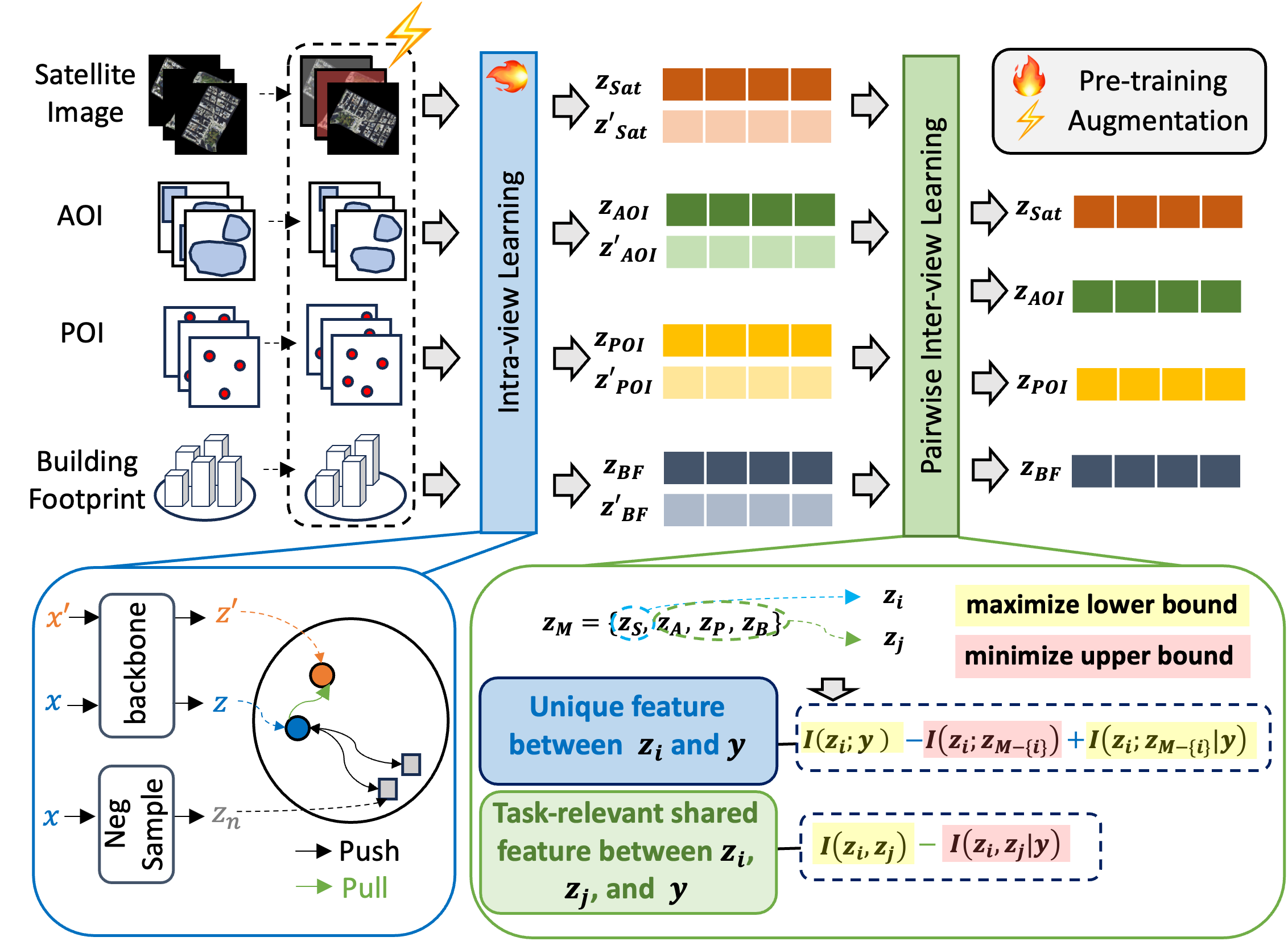}
    \vspace{0.5em}
    \caption{\modelname consists of two goals: (a) intra-view learning, which captures each modality's representation by comparing other regions, and (b) pairwise inter-view learning, which extracts task-relevant unique and shared information across modalities. \light refers to an augmentation and \fire represents pretraining each modality's backbone.}
    \label{cookie_method}
\end{figure}

Figure~\ref{cookie_method} illustrates \modelnamex's architecture, where $m$ encoders transform modalities into embeddings. \modelname first pretrains each encoder independently to enhance region distinctiveness, then jointly trains the embeddings using inter-view learning. Building on FactorCL~\cite{factorcl}, we extend its approach to support more than two modalities. Further, we propose a pairwise inter-view learning strategy to capture both shared information between modalities and unique information within each modality. The refined embeddings serve as comprehensive region representations, which act as covariates for predicting various downstream tasks.

\subsection{Pretraining via Intra-view learning}
\modelname first pretrains the encoders for each modality using intra-view contrastive learning, which encourages the embeddings to be distinctive across regions (shown in intra-view learning in Figure~\ref{cookie_method}).
We use the Noise Contrastive Estimation (NCE) loss~\cite{tkde, oord2018representation}, treating an augmented view of the same region as a positive sample and other regions as negative samples. 

\textbf{POIs and AOIs.} For POI and AOI data, where each element describes an environmental characteristic (e.g., the number of restaurants, the area of parks), we use a multilayer perceptron (MLP) as the encoder. To create positive samples for intra-view contrastive learning, we apply three data augmentation techniques~\cite{tkde}: random addition, random removal, and random replacement with a $0.1$ probability. 

\textbf{Satellite Imagery.} We use a pretrained ResNet50 as the encoder to extract features from satellite images. For data augmentation, we use random flipping, color jittering, and Gaussian noise, following~\cite{simclr} while excluding random cropping as in~\cite{factorcl}.

\textbf{Building Footprints.}
To preserve the two-dimensional shapes of buildings, we first use ResNet18 to encode individual raw building footprints into visual features~\cite{li2023urban}. We then use a MLP as region feature encoder by aggregating the visual features within each region to generate region-level feature vectors.
For region-level data augmentation, we randomly drop 10\% of the building features to create a positive pair.


\textbf{Intra-view Loss} is defined as:
\begin{align}
\mathcal{L}_{intra} &= I_{NCE}(z_i, z_j)\nonumber \\
&= \sum_{i=1}^m I_{NCE}(z_i; z_i^{+}, \{z_{i,j}^{-}\}_{j=1}^N) \nonumber \\
&= \sum_{i=1}^m-\log \frac{\exp(z_i \cdot z_i^{+} / \tau)}{\exp(z_i \cdot z_i^{+} / \tau) + \sum_{j=1}^{N} \exp(z_i \cdot z_{i,j}^{-} / \tau)} \label{eq:L_intra}
\end{align}
where $I_{NCE}$ denotes NCE loss~\cite{oord2018representation}, which estimates a lower bound on the mutual information between the anchor embedding $z_i$ and its positive sample $z_i^+$, given a set of negative samples $\{z_{i,j}^{-}\}_{j=1}^N$. Here, $z_i$ is the embedding of a region obtained from a modality-specific encoder, and $z_i^+$ is an augmentation of $z_i$ generated using each augmentation method varying by modality. The negatives $z_{i,j}^-$ are embeddings from other regions. This process encourages \modelname to distinguish between other regions by maximizing the similarity between positive pairs while minimizing similarity to negative samples.

\subsection{Pairwise Inter-view learning}
Inter-view learning aims to align information from multiple modalities for a region. We build task-relevant unique and shared information across modalities via pairwise inter-view learning.
\subsubsection{Preliminary: Mutual Information}
Mutual information (MI) quantifies the amount of information shared between two variables (representations). Formally, the MI between variables \( X \) and \( Y \) is defined as 
\[I(X;Y) = \mathbb{E}_{p(x,y)}\left[\log\frac{p(x,y)}{p(x)p(y)}\right]. \]
Intuitively, a high mutual information value indicates that knowing one variable significantly reduces uncertainty about the other. In our methodology, we utilize variants of MI, such as conditional mutual information \( I(X;Y\mid Z) \) (where $X, Y, Z$ are each modality and $I$ is MI), to distinguish unique and shared information among multiple data modalities. Practically, since the exact computation of mutual information is often infeasible~\cite{clubloss}, we leverage InfoNCE~\cite{oord2018representation} and CLUB~\cite{clubloss}, which provide reliable lower and upper bounds to estimate MI. These estimators translate MI maximization into tractable optimization objectives that guide representation learning.

\subsubsection{Preliminary: FactorCL}
FactorCL~\cite{factorcl} is an information factorization approach designed to disentangle task-relevant shared and unique information across two modality-specific representations \( \{z_1, z_2\} \), while aligning representations. Since mutual information (MI) between high-dimensional embeddings is generally intractable to compute directly, FactorCL adopts tractable bounds: InfoNCE~\cite{oord2018representation} as a lower bound, and CLUB~\cite{clubloss} as an upper bound.

To capture \textit{shared information}, FactorCL maximizes the lower bound of mutual information between modalities via \( I_{NCE}(z_1, z_2) \) and removes task-irrelevant information by minimizing the conditional mutual information (CMI) upper bound \( -I_{CLUB}(z_1, z_2 \mid y) \), where \( y \) denotes task-relevant semantics. For \textit{unique information}, FactorCL maximizes \( I_{NCE}(z_1, y) \) to preserve modality-specific information, and minimizes \( -\left( I_{CLUB}(z_1, z_2) - I_{NCE}(z_1; z_2 \mid y) \right) \) to remove task-irrelevant information.

\( I(z_1; z_2 \mid y) \) is estimated using the conditional InfoNCE loss:

\begin{align}
I_{NCE}(z_1; z_2 \mid y) \nonumber \\
&\!\!\!\!\!\!\!\!\!\!\!\!\!\!\!\!\!\!\!\!\!\!\!\!\!\!\!\!\!\!\!\!\!\!\!\!\!\!\!\!\!\!\!\!\!= \!\mathbb{E}_{y \sim p(y)}\!\! \left[
    \mathbb{E}_{\substack{(z_1, z_2^{+}) \sim p(z_1, z_2 \mid y) \\
                          \{z_2^{-j}\}_{j=1}^N \sim p(z_2 \mid y)}} 
    \left[ 
        \frac{
            W(z_1, z_2^{+})
        }{
            W(z_1, z_2^{+}) + \sum_{j=1}^{N} W(z_1, z_2^{-j})
        }
    \right]
\right],
\label{eq:infoNCE}
\end{align}
where \( z_1 \) and \( z_2 \) are representations from two modalities, and \( W(z_1, z_2) \) denotes a similarity score, computed as \( \exp(f(\cdot)) \) (e.g., $f(\cdot)$ is a MLP over concatenated embeddings). The positive sample \( z_2^{+} \) shares the same semantic label \( y \) with \( z_1 \), while negatives \( \{z_2^{-j}\} \) are drawn from other samples conditioned on the same conditional distribution \( p(z_2 \mid y) \). This contrastive formulation provides a scalar loss via log-softmax over one positive and multiple negatives, forming a tractable lower bound on \( I(z_1; z_2 \mid y) \) in~\cite{factorcl}.

Similarly, \( I_{CLUB} \) estimates an upper bound on task-irrelevant information by modeling CMI as:

\begin{align}
I_{CLUB}(z_1; z_2 \mid y) \nonumber\\
&\!\!\!\!\!\!\!\!\!\!\!\!\!\!\!\!\!\!\!\!\!\!\!\!\!\!\!\!\!\!\!\!\!\!\!\!\!\!\!\!\!\!\!\!\!\!\!= \mathbb{E}_{y \sim p(y)} \!\!\left[
    \mathbb{E}_{(z_1, z_2^+) \sim p(z_1, z_2 \mid y)} f(z_1, z_2^+) 
    - \mathbb{E}_{\substack{z_1 \sim p(z_1 \mid y) \\ z_2^- \sim p(z_2 \mid y)}} f(z_1, z_2^-)
\!\right]\!,
\label{eq:club}
\end{align}
where \( f(z_1, z_2) \) is the same learnable scoring function as in InfoNCE from Equation~\eqref{eq:infoNCE}. The first expectation is over positively paired representations \( (z_1, z_2^+) \sim p(z_1, z_2 \mid y) \). The second expectation is taken over representations sampled independently from the marginals $p(z_1 \mid y)$ and $p(z_2 \mid y)$, which removes cross-modal correlation.
Minimizing this difference penalizes redundant statistical dependencies between $z_1$ and $z_2$ that persist even when conditioned on $y$. Therefore, InfoNCE in Equation ~\eqref{eq:infoNCE} aligns semantically related modality pairs and CLUB in Equation~\eqref{eq:club} encourages disentanglement by reducing task-irrelevant information.


\subsubsection{Generalized FactorCL (GFactorCL) using $\geq$ 3 modalities}
Following the estimation strategies used in \( I_{NCE} \) and \( I_{CLUB} \), we build a generalized FactorCL (GFactorCL) to handle three modalities \( \{x_1, x_2, x_3\} \). GFactorCL aims to learn three types of task-relevant information: 
(1) \textbf{unique information} specific to each modality, represented as \( Z_{U_1} \) for \( x_1 \), \( Z_{U_2} \) for \( x_2 \), and \( Z_{U_3} \) for \( x_3 \),
(2) \textbf{conditional pairwise shared information} between two modalities given a third, \( Z_{S_{12\mid3}} \) for \( x_1 \) and \( x_2 \), \( Z_{S_{13\mid2}} \) for \( x_1 \) and \( x_3 \), and \( Z_{S_{23\mid1}} \) for \( x_2 \) and \( x_3 \)\footnote{We denote representation as $Z$ and scalar computation for obtaining lower and upper bound to estimate mutual information as $U$ (i.e., unique information) and $S$ (i.e., shared information).}. 
(3) \textbf{high-order shared information} among all three modalities (e.g., \( Z_{S_{123}} \)). 
The unique information of modality \( z_1 \) is estimated as:
\begin{align}
    U_1 &= I(z_1; y \mid z_{\mathcal{M} \setminus \{z_1\}}) \nonumber \\
    &= I_{NCE}(z_1; y) - I_{CLUB}(z_1; z_{\mathcal{M} \setminus \{z_1\}}) 
    + I_{NCE}(z_1; z_{\mathcal{M} \setminus \{z_1\}} \mid y),
    \label{gfactorcl:u}
\end{align}
where \( I(\cdot) \) represents the mutual information, $y$ is task-relevant semantics, and \( z_{\mathcal{M} \setminus \{z_1\}} \) refers to all modalities excluding \( z_i \). We consider all possible unique information in the pairwise approach. For example, we consider conditions like both \( I(z_1; y \mid z_2) \) and \( I(z_1; y \mid z_3) \) when \( \mathcal{M} = 3 \) to learn \( U_1 \) conditioning on $z_2$ and $z_3$, respectively, for ensuring comprehensive representation. Similarly, $U_2$ and $U_3$ follows Equation~\eqref{gfactorcl:u}. We maximize the lower bound of mutual information using $I_{NCE}$ and minimize task-irrelevant dependencies through $I_{CLUB}$.
Similarly, the conditional shared information between two modalities given a third is:
\begin{align}
    S_{12|3} &= I(z_1; z_2; y \mid z_3) = I_{NCE}(z_1; z_2 \mid z_3) 
    - I_{CLUB}(z_1; z_2 \mid z_3, y),
    \label{gfactorcl:s}
\end{align}
where $I_{NCE}$ first captures shared information between $z_1$ and $z_2$ given conditioned on $z_3$ then removes task-irrelevant information via $I_{CLUB}(z_1; z_2\!\!  \mid\!\! z_3, y)$. Similarly, $S_{13\mid2}$ and $S_{23\mid1}$ follow Equation~\eqref{gfactorcl:s}.

We also define the high-order shared information among all three modalities using the concept of \textit{interaction information}~\cite{mcgill1954multivariate}, which captures how the mutual dependence among variables changes when conditioned on a fourth variable. In our case, we use interaction information to quantifies the change between the three modalities \( \{z_1, z_2, z_3\} \) when the task label \( y \) is observed:

\begin{align}
    S_{123} & = I(z_1; z_2; z_3; y) \nonumber \\
    & =  I(z_1;z_2;z_3) - I(z_1;z_2;z_3\mid y) \nonumber \\
    &= I(z_1, z_2) - I(z_1;z_2\mid z_3)
    - \left(I(z_1;z_2\mid y) - I(z_1;z_2\mid z_3,y)\right) \nonumber \\
    &= I_{NCE}(z_1; z_2) - I_{CLUB}(z_1;z_2\mid z_3) \nonumber \\
    & \quad\quad - I_{CLUB}(z_1;z_2\mid y) + I_{NCE}(z_1;z_2\mid z_3,y),
    \label{gfactorcl:sall}
\end{align}
where each term is estimated following the strategy used in Equations~\eqref{gfactorcl:u} and~\eqref{gfactorcl:s}. This formulation extends to capture dependencies that arise only through the joint interaction of all three modalities with the task.

When \( \mathcal{M} = 3 \), GFactorCL learns seven objectives: three for \textbf{unique information} (\( U_1, U_2, U_3 \)), three for \textbf{conditional pairwise} shared information (\( S_{12\mid3}, S_{13\mid2}, S_{23\mid1} \)), and one for \textbf{high-order} shared information across all modalities (\( S_{123} \)). 
However, as the number of modalities increases, modeling high-order dependencies leads to a combinatorial growth in the number of learning objectives and requires the computation of CMI for every shared information. 

\paragraph{\textbf{Loss function}}
We define three loss components for GFactorCL inter-view learning: one for capturing \textit{unique information} per modality, one for \textit{conditional shared information}, and one for \textit{high-order shared information} across all modalities.
\begin{align}
    \mathcal{L}_{\text{inter}_u} 
    &= \sum_{i=1}^{\mathcal{M}} U_i \nonumber \\
    &= \sum_{i \neq j} \left( I_{NCE}(z_i; y) - I_{CLUB}(z_i; z_j) + I_{NCE}(z_i; z_j \mid y) \right), \label{eq:L_u} \\
    \mathcal{L}_{\text{inter}_s} 
    &= \sum_{\substack{i \neq j \\ k \in \mathcal{M} \setminus \{i,j\}}} \left( 
        I_{NCE}(z_i; z_j \mid z_k) - I_{CLUB}(z_i; z_j \mid z_k, y)
    \right), \label{eq:L_s} \\
    \mathcal{L}_{\text{inter}_h} 
    &= S_{ijk} \nonumber \\
    &= I_{NCE}(z_i; z_j) - I_{CLUB}(z_i; z_j \mid z_k) \\
    & \quad \quad - I_{CLUB}(z_i; z_j \mid y) + I_{NCE}(z_i; z_j \mid z_k, y), \label{eq:L_h}
\end{align}
where \( \mathcal{L}_{\text{inter}_u} \) captures \textit{unique information} for each modality \( z_i \) in Equation~\eqref{gfactorcl:u}. 
The term \( \mathcal{L}_{\text{inter}_s} \) corresponds to the conditional shared information \( S_{ij \mid k} = I(z_i; z_j; y \mid z_k) \) across all triplets of modalities, as defined in Equation~\eqref{gfactorcl:s}.
Finally, \( \mathcal{L}_{\text{inter}_h} \) captures high-order shared information \( S_{ijk} \), based on the interaction information across three modalities conditioned on the task label \( y \), as shown in Equation~\eqref{gfactorcl:sall}. This loss measures dependency structures that only emerge when considering all three modalities jointly.
Together, these losses allow GFactorCL to disentangle and learn both task-relevant and task-irrelevant information of multimodal interactions, while maintaining tractability through lower and upper bound estimators~\cite{factorcl}.

\subsubsection{Pairwise inter-view learning.}
We propose a \emph{pairwise inter-view learning} approach to capture high-order dependencies, as illustrated in Figure~\ref{cookie_method}. By modeling interactions across all modality pairs, \modelname extracts both task-relevant shared and unique information from each modality.
We define the unique information of a representation \(z_i\) of modality \(x_i\) and the shared information between \(z_i\) and \(z_j\) by:
\begin{align}
    U_i & = I(z_i; y \mid z_{\mathcal{M} \setminus \{z_i\}}), \label{method:U} \\
    S_{ij} & = I(z_i; z_j; y) = I_{NCE}(z_i; z_j) - I_{CLUB}(z_i; z_j \mid y), \label{method:S} 
\end{align}
where \( {\mathcal{M} \setminus \{z_i\}} \) denotes all modalities excluding \( z_i \), and \( \{(z_i, z_j) \mid i \neq j, i,j \in \{1, ..., \mathcal{M}\} \} \) represents all distinct modality pairs in the set.

Note that InfoNCE is a lower bound on mutual information~\cite{oord2018representation}, and its use in Equation~\eqref{method:S} does not imply equality with true MI. Rather, we adopt \( I_{NCE} \) as an estimation for mutual information under the assumption that maximizing the lower bound generally increases the true value. While \( I_{NCE} \) may not exactly match \( I(z_i; z_j) \), prior work~\cite{factorcl} demonstrates that using \( I_{NCE} \) and \( I_{CLUB} \) is a useful and tractable proxy, particularly when comparing modality pairs.

\paragraph{\textbf{Theoretical Analysis.}}
We compare the shared information captured by \modelnamex's pairwise inter-view learning with GFactorCL\footnote{We adapt FactorCL when capturing unique information in inter-view learning.}. When $\mathcal{M} = 3$, \modelname represents shared information using unconditioned pariwise terms, $S_{12}, S_{13},S_{23}$, while GFactorCL explicitly factorizes shared information using CMI, expressed as $S_{12\mid3}, S_{13\mid2}, S_{23\mid1}, S_{123}$.
GFactorCL separates shared information into conditional pairwise and high-order information, whereas \modelname does not condition each pair on the third modality, potentially capturing overlapping information across pairs. This lease to the inclusion relationship:
\begin{align}
    \{S_{12}, S_{13}, S_{23}\} \supseteq
    \{S_{12 \mid 3}, S_{13 \mid 2}, S_{23 \mid 1}, S_{123}\}\label{eq:proof},
\end{align}
which indicates that pairwise inter-view learning in \modelname may redundantly account for parts of the high-order shared information. Thus, explicit modeling of \( S_{123} \) may not always be necessary when such redundancies are already embedded within unconditioned pairs.
Equation~\eqref{eq:proof} can be written by the interaction information concept~\cite{mcgill1954multivariate} following $I(z_1; ... ;z_{n+1})$ $= I(z_1; ... ;z_{n}) - I(z_1; ... ;z_{n}\mid z_{n+1})$. $\{S_{12}, S_{13}, S_{23}\}$ can be written by:

\begin{align}
    \{I(z_1; z_2), I(z_1; z_3), I(z_2; z_3)\} \nonumber \\    &\!\!\!\!\!\!\!\!\!\!\!\!\!\!\!\!\!\!\!\!\!\!\!\!\!\!\!\!\!\!\!\!\!\!\!\!\!\!\!\!\!\!\!\!\!\!\!\!\!\!\!\!\!\!\!\!\!\!\!\!\!\!\!\!\!\!\!\!\!\!\!\!\!\!\!\!\!\!\!\! = \{[I(z_1;z_2;z_3)+ I(z_1;z_2|z_3)], [I(z_1;z_2;z_3) + I(z_1;z_3|z_2)], \nonumber\\
&\!\!\!\!\!\!\!\!\!\!\!\!\!\!\!\!\!\!\!\!\!\!\!\!\!\!\!\!\!\!\!\!\!\!\!\!\!\!\!\!\!\!\!\!\!\!\!\!\!\!\!\!\!\!\!\!\!\!\!\!\!\!\!\!\!\!\! [I(z_1;z_2;z_3) + I(z_2;z_3|z_1)] \} \nonumber\\ 
&\!\!\!\!\!\!\!\!\!\!\!\!\!\!\!\!\!\!\!\!\!\!\!\!\!\!\!\!\!\!\!\!\!\!\!\!\!\!\!\!\!\!\!\!\!\!\!\!\!\!\!\!\!\!\!\!\!\!\!\!\!\!\!\!\!\!\!\!\!\!\!\!\!\!\!\!\!\!\!\! = \{3 \times I(z_1;z_2;z_3) ,  I(z_1;z_2|z_3), I(z_1;z_3|z_2), I(z_2;z_3|z_3)\} .
\end{align}
Therefore, 
\begin{align}
    \{3 \times I(z_1;z_2;z_3) ,  I(z_1;z_2|z_3), I(z_1;z_3|z_2), I(z_2;z_3|z_3)\} \\
    \supseteq  \{I(z_1;z_2|z_3),I(z_1;z_3|z_2),I(z_2;z_3|z_1),I(z_1;z_2;z_3) \}. \label{eq:cookie_proof} 
\end{align}

Additionally, the pairwise approach in \modelname minimizes the number of learning objectives.
For example, when $\mathcal{M} = 3$, our method requires six objectives ($U_1$, $U_2$, $U_3$, $S_{12}$, $S_{13}$, $S_{23}$), while GFactorCL results in seven objectives, adding the high-order term $S_{123}$. The difference becomes more significant as the number of modalities increases. With four modalities, our approach uses $10$ objectives ($U_1$, $U_2$, $U_3$, $U_4$, $S_{12}$, $S_{13}$, $S_{14}$, $S_{23}$, $S_{24}$, $S_{34}$) compared to GFactorCL’s $15$, which includes additional high-order terms ($S_{123|4}$, $S_{124|3}$ ,$S_{134|2}$, $S_{234|1}$, $S_{1234}$).
Moreover, decomposing each objective into $I_{NCE}$ and $I_{CLUB}$ terms results in a quadratic scaling with the number of modalities, $O(m^2)$, because our pairwise approach considers only interactions between modality pairs. In contrast, GFactorCL scales exponentially, $O(2^m)$, as it accounts for all possible combinations of modalities, including high-order dependencies. 

While \modelname captures the high-order relationships multiple times (see Equation~\eqref{eq:cookie_proof}), \textbf{this repetition does not increase computational complexity}. Further, the machine learning-based predictor can filter out redundant information from the learned representation.

\textbf{Loss function.}
We define two loss components for pairwise inter-view learning, targeting the estimation of unique and shared task-relevant information across modality pairs:
\begin{align}
    \mathcal{L}_{\text{inter}_u} 
    &= \sum_{i=1}^{\mathcal{M}} U_i \nonumber \\
    &= \sum_{i \neq j} \left( I_{NCE}(z_i; y) - I_{CLUB}(z_i; z_j) + I_{NCE}(z_i; z_j \mid y) \right), \label{eq:L_u} \\
    \mathcal{L}_{\text{inter}_s} 
    &= \sum_{i=1}^{\mathcal{M}} \sum_{j \in \mathcal{M} \setminus \{i\}} S_{ij} \nonumber \\
    &= \sum_{i \neq j} \left( I_{NCE}(z_i; z_j) - I_{CLUB}(z_i; z_j \mid y) \right), \label{eq:L_s}
\end{align}
where \( \mathcal{L}_{\text{inter}_u} \) corresponds to the estimation of \textit{unique information} for each modality \( z_i \), following the decomposition in Equation~\eqref{gfactorcl:u}. 
This loss is computed for all modality pairs \( (z_i, z_j) \), enabling estimation of \( U_i \) using only pairwise interactions.
Similarly, \( \mathcal{L}_{\text{inter}_s} \) estimates the \textit{shared information} between modality pairs \( (z_i, z_j) \), as defined in Equation~\eqref{method:S}. 
Together, these loss functions implement pairwise inter-view learning that approximates the shared and unique information components across modalities, using tractable lower and upper bounds as proposed in FactorCL~\cite{factorcl}.


\subsubsection{Optimal Augmentation}\label{bg:aug}
In self-supervised learning, where labels are absent, it is crucial to design augmentations that capture task-relevant semantics. As demonstrated in FactorCL~\cite{factorcl}, they use the \textit{optimal unimodal augmentation} assumption to capture task-relevant information without relying on labels. Under this assumption, an augmented version of the modality $x$, denoted as $x^\prime$, serves as a substitute for the label $y$. This means that when all information shared between $x$ and $x^\prime$ is task-relevant, the augmentation preserves task-relevant information while changing task-irrelevant information. 
$I(x;y) = I(x;x^\prime)$ ensures that $x^{\prime}$ retains the same semantic to $y$. If an augmentation discards essential semantic features (e.g., by aggressive cropping), then the shared information decreases, resulting in $I(x; x^{\prime}) < I(x; y)$. 

Hence, FactorCL proposes \textit{optimal unimodal augmentation} that builds on the idea of the \textit{optimal representation of a task}~\cite{tian2020makes}. This augmentation strategy is built upon that an encoder should produce a \textit{minimal sufficient statistic}. This statistic retains all label-relevant information while discarding irrelevant details. By treating $x^{\prime}$ (or its encoded representation $z^{\prime}$) as a valid proxy for $y$, \modelname can learn a representation $z$ that is both sufficient (preserving all task-relevant information) and minimal (excluding unnecessary information) for downstream prediction tasks.


\subsection{\modelname objectives}
In the final stage, \modelname integrates intra-view and pairwise inter-view learning losses into a joint objective function:
\begin{align}
    \mathcal{L} 
    &= \alpha \mathcal{L}_{intra} + \mathcal{L}_{inter_s} + \mathcal{L}_{inter_u}
\end{align}
where $\mathcal{L}_{intra}$ is the intra-view learning loss in Equation~\eqref{eq:L_intra}, and $\mathcal{L}_{inter_s}$ and $\mathcal{L}_{inter_u}$ denote the shared and unique inter-view learning losses in Equation~\eqref{eq:L_u} and~\eqref{eq:L_s}. We include $\alpha$ to match the scale with $\mathcal{L}_{inter_s}$ and $\mathcal{L}_{inter_u}$. 
After training \modelnamex, we use the learned region embeddings as covariates, applying a random forest regressor for predictions and an MLP for a classification. 
\section{Experiments}
We evaluate \modelname by using the learned embeddings to predict population density, crime rates, land use functions, and greenness score. The first three tasks are common downstream tasks in RRL~\cite{li2023urban,deng2024novel}, and street-level greenness score is an important indicator of a region’s environmental quality and public health~\cite{dronova2018spatio,green}. We test four downstream tasks in New York City (NYC) and extend the greenness score prediction task to Delhi, India, to assess the model's performance in two cities with varying geographic partition sizes. 

\subsection{Datasets} 
The total geographic regions in NYC is 2317 across five boroughs and in Delhi is 000 tracts across the city. We sample data for each geographic unit. 
\textbf{POIs ($\mathcal{P}$) and AOIs ($\mathcal{A}$).}
We extract POI and AOI features from OpenStreetMap (OSM) and aggregate them into 27 predefined categories for POI and 34 for AOI~\cite{zhao2023learning}.
For OSM types that do not correspond to the existing categories, we define additional categories to ensure comprehensive coverage, such as region boundaries.

\textbf{Satellite Imagery ($\mathcal{S}$).}\label{exp:sat}
For NYC, we collect satellite imagery from NAIP\footnote{\url{https://naip-usdaonline.hub.arcgis.com/}} with a 1-meter resolution (2021) and for Delhi from PLANET\footnote{\url{https://www.planet.com/}} with a 3-meter resolution (2023). 
To standardize the image dimension ($1024\times1024$ pixels), we apply three methods: (1) if an image is larger than the standard size on both sides, we divide it into patches ($256\times256$), and randomly select (at most 16 patches) to merge into a single image to guarantee spatial coverage; (2) if an image has only one side larger than 1024, we crop the long side to fit the standard size; and (3) for the rest, we pad the images to meet the desired dimension.

\textbf{Building Footprints ($\mathcal{B}$).} 
We obtain 2D polygonal building footprint data from OSM for NYC and Delhi. 

\textbf{Evaluation Datasets.}
For NYC, we collect the following datasets: population density from WorldPoP\footnote{\url{https://hub.worldpop.org/geodata/listing?id=77}}, land use data from MapPLUTO\footnote{\url{https://www.nyc.gov/site/planning/data-maps/open-data/dwn-pluto-mappluto.page}}, crime data from Open Data\footnote{\url{https://opendata.cityofnewyork.us/}}, and greenness score from Treepedia~\cite{green}. 
For Delhi, we generate street-level greenness score following~\cite{green}. We query Google Street View images at road intersections, covering four directions. We then apply semantic segmentation using OneFormer~\cite{oneformer}, fine-tuned with the Cityscapes dataset~\cite{cityscapes}. The greenness score at each intersection is calculated as $\text{Greenness} =\frac{\sum_{i=1}^4Area_{g}}{\sum_{i=1}^4Area_{a}}$, where $Area_{g}$ represents the number of green pixels (i.e., vegetation labels) and $Area_{a}$ is the total number of pixels across the four images.
Finally, we average the greenness score within each ward to obtain the ward-level greenness score. 

\subsection{Baselines \& Training Details}
We compare \modelname to seven state-of-the-art methods, categorized into two groups: inter-view only methods (MVURE~\cite{mvure}, KnowCL~\cite{liu2023knowledge}, FactorCL~\cite{factorcl}) and intra- \& inter-view learning methods (MGFN~\cite{wu2022multi_mgfn}, RegionDCL~\cite{li2023urban}, Urban2Vec~\cite{wang2020urban2vec}, ReMVC~\cite{tkde}).

During pretraining, we set the embedding size to 512. The final embedding is obtained by concatenating the learned representations from all learning objectives. We use the Adam optimizer with a learning rate of 0.0001.

\subsection{Evaluation \& Downstream Test Settings}
We evaluate population density, crime rate, and greenness score using mean absolute error ($MAE$), root mean square error ($RMSE$), and R-square ($R^2$)~\cite{li2023urban}. 
For the 5-class land use classification (Residential, Industrial, Commercial, Open Space, and Others.), we use L1 distance, KL-divergence, and cosine similarity~\cite{li2023urban}.

For the three regression tasks, we randomly split the regions into five folds (80\% training, 20\% testing) and use a random forest model for prediction. For the classification task, we also test the same five folds using a 2-layer MLP with a hidden dimension of 512. The MLP is optimized using KL-divergence loss for 300 epochs, following~\cite{li2023urban}. 



\section{Results \& Discussion}

\begin{table*}[h]
\centering
\caption{Comparison of \modelname with baseline models for population density and crime rate prediction in NYC. $\downarrow$ indicates that a lower value is better, while $\uparrow$ indicates the opposite. The best results are bolded, and the second best results are underlined. Modalities include POI ($\mathcal{P}$), AOI ($\mathcal{A}$), satellite imagery ($\mathcal{S}$), building footprints ($\mathcal{B}$), mobility ($\mathcal{MB}$) and  knowledge graph ($\mathcal{KG}$).}
\begin{adjustbox}{width=1.0\textwidth,center}
\begin{tabular}{cccccccc}
\toprule
\multicolumn{1}{c}{} & \multicolumn{1}{c}{} & \multicolumn{3}{c}{\textbf{Population Density}} & \multicolumn{3}{c}{\textbf{Crime Rate}}\\
\cmidrule(rl){3-5} \cmidrule(rl){6-8} 
\textbf{Methods} & \textbf{Modalities} & {MAE} $\downarrow$ & {RMSE} $\downarrow$ & {R2} $\uparrow$ & {MAE} $\downarrow$ & {RMSE} $\downarrow$ & {R2}$\uparrow$\\
\midrule
MGFN & $\mathcal{MB}$ & 6754.58$\pm$201.13 & 8989.14$\pm$312.71 & 0.331$\pm$0.050 & 116.08$\pm$4.40 & 161.94$\pm$9.86 & 0.231$\pm$0.043   \\
MVURE & $\mathcal{B}$ + $\mathcal{P}$ + $\mathcal{MB}$ & 5295.00$\pm$101.72 & 7065.08$\pm$122.91 & 0.545$\pm$0.014    & 108.60$\pm$5.01 & 151.87$\pm$13.46 & 0.278$\pm$0.043 \\ 
RegionDCL & $\mathcal{B}$ + $\mathcal{P}$ + $\mathcal{A}$ & 4891.96$\pm$103.11 & 6890.84$\pm$164.84 & 0.568$\pm$0.013     & 106.00$\pm$2.36 & 163.71$\pm$15.17 & 0.290$\pm$0.034 \\
%
Urban2Vec & $\mathcal{S}$ + $\mathcal{P}$ + $\mathcal{A}$ & 4760.06$\pm$191.45 & 6497.80$\pm$395.89 & 0.614$\pm$0.053 & 104.95$\pm$3.49 & 159.28$\pm$12.24 & 0.326$\pm$0.040 \\ 
ReMVC & $\mathcal{S}$ + $\mathcal{P}$ + $\mathcal{A}$ & 7321.98$\pm$122.16 & 9189.99$\pm$196.61 & 0.190$\pm$0.016 & 119.68$\pm$3.28 & 166.94$\pm$12.16 & 0.182$\pm$0.039 \\ 
GFactorCL & $\mathcal{S}$ + $\mathcal{P}$ + $\mathcal{A}$ & 4991.93$\pm$188.62 & 6792.49$\pm$336.46 & 0.567$\pm$0.030 & 111.68$\pm$3.52 & 165.62$\pm$12.09 & 0.270$\pm$0.041  \\
 \midrule 
\modelname  & $\mathcal{B}$ + $\mathcal{P}$ + $\mathcal{A}$ & 4562.71$\pm$43.51 & 6316.58$\pm$89.46 & 0.627$\pm$0.009 & 102.28$\pm$4.32 & \textbf{147.35$\pm$13.55} & \underline{0.346$\pm$0.030}  \\
\modelname& $\mathcal{S}$ + $\mathcal{P}$ + $\mathcal{A}$  & 4627.80$\pm$163.49 & 6306.03$\pm$296.37 & 0.630$\pm$0.033 & 105.47$\pm$3.17 & 149.54$\pm$11.40 & 0.322$\pm$0.030   \\ 
\midrule
GFactorCL & $\mathcal{S}$ + $\mathcal{B}$ + $\mathcal{P}$ + $\mathcal{A}$ & \underline{4327.07$\pm$103.80} & \underline{6079.85$\pm$229.24} & \underline{0.664$\pm$0.014} & \underline{101.32$\pm$1.37} & 152.76$\pm$10.58 & 0.329$\pm$0.013 \\ 
\modelname & $\mathcal{S}$ + $\mathcal{B}$ + $\mathcal{P}$ + $\mathcal{A}$ & \textbf{4102.23$\pm$124.62} & \textbf{5584.12$\pm$221.17} &  \textbf{0.698$\pm$0.017} & \textbf{99.20$\pm$1.65}& \underline{148.52$\pm$11.37} & \textbf{0.377$\pm$0.017} \\ 
\midrule \midrule 
KnowCL (log) & $\mathcal{S}$ + $\mathcal{KG}$ & - & 0.612 & 0.153 & - & 0.622 & 0.536  \\ 
\modelname (log) & $\mathcal{S}$ + $\mathcal{B}$ + $\mathcal{P}$ + $\mathcal{A}$  & - &  0.399 & 0.557 & - & 0.621 & 0.534 \\ \bottomrule
\end{tabular}
\label{table_result_1}
\end{adjustbox}
\end{table*}

\begin{table*}[h]
\centering
\caption{Comparison of \modelname with baseline models for greenness score and land use classification in NYC.}
\begin{adjustbox}{width=1.0\textwidth,center}
\begin{tabular}{cccccccc}
\toprule
\multicolumn{1}{c}{} & \multicolumn{1}{c}{} & \multicolumn{3}{c}{\textbf{Greenness Score}} & \multicolumn{3}{c}{\textbf{Land Use}}\\
\cmidrule(rl){3-5} \cmidrule(rl){6-8} 
\textbf{Methods} & \textbf{Modalities} & {MAE}$\downarrow$ & {RMSE}$\downarrow$ & {R2}$\uparrow$ & {L1}$\downarrow$ & {KL}$\downarrow$ & {Cosine}$\uparrow$\\
\midrule
MGFN & $\mathcal{MB}$ & 3.949$\pm$0.187 & 5.032$\pm$0.255 & 0.094$\pm$0.023 & 0.530$\pm$0.012 & 0.334$\pm$0.011 & 0.875$\pm$0.006\\
MVURE & $ \mathcal{B} + \mathcal{P} +\mathcal{MB} $ & 3.723$\pm$0.201 & 4.875$\pm$0.259 & 0.183$\pm$0.039 & 0.526$\pm$0.008 & 0.337$\pm$0.011 & 0.872$\pm$0.006 \\ 
RegionDCL &$\mathcal{B} + \mathcal{P} + \mathcal{A}$& 3.481$\pm$0.176 & 4.574$\pm$0.254 & 0.290$\pm$0.032 & 0.488$\pm$0.005 & 0.273$\pm$0.007 & 0.893$\pm$0.005 
\\ 
Urban2Vec &$\mathcal{S}$ + $\mathcal{P}$ + $\mathcal{A}$&  \underline{3.373$\pm$0.163} & 4.423$\pm$0.239 & \textbf{0.337$\pm$0.028} & 0.421$\pm$0.007 & \textbf{0.223$\pm$0.007} & \textbf{0.915$\pm$0.006} \\
ReMVC & $\mathcal{S}$ + $\mathcal{P}$ + $\mathcal{A}$ & 4.051$\pm$0.192 & 5.151$\pm$0.240 & 0.068$\pm$0.026 & 0.608$\pm$0.029 & 0.475$\pm$0.027 & 0.831$\pm$0.012 \\ 
GFactorCL  & $\mathcal{S}$ + $\mathcal{P}$ + $\mathcal{A}$ & 3.690$\pm$0.150 & 4.766$\pm$0.184 & 0.208$\pm$0.036 & 0.441$\pm$0.013 & 0.271$\pm$0.015 & 0.900$\pm$0.005 \\
\midrule
\modelname  & $\mathcal{B}$ + $\mathcal{P}$ + $\mathcal{A}$ & 3.444$\pm$0.175 & 4.371$\pm$0.246 & 0.292$\pm$0.011 & 0.442$\pm$0.010 & 0.267$\pm$0.080 & 0.901$\pm$0.003 
\\  
\modelname & $\mathcal{S}$ + $\mathcal{P}$ + $\mathcal{A}$  & 3.573$\pm$0.157 & 4.583$\pm$0.221 & 0.251$\pm$0.017 & 0.424$\pm$0.010 & 0.236$\pm$0.060 & \underline{0.913$\pm$0.003} \\ \midrule
GFactorCL & $\mathcal{S}$ + $\mathcal{B}$ + $\mathcal{P}$ + $\mathcal{A}$ & 3.462$\pm$0.200 & \underline{4.323$\pm$0.288} & 0.296$\pm$0.017 & \textbf{0.398$\pm$0.01} & 0.244$\pm$0.011 & 0.912$\pm$0.006
  \\ 
\modelname & $\mathcal{S}$ + $\mathcal{B}$ + $\mathcal{P}$ + $\mathcal{A}$ & \textbf{3.368$\pm$0.179} & \textbf{4.300$\pm$0.246} & \underline{0.325$\pm$0.014} & \underline{0.411$\pm$0.016} & \underline{0.226$\pm$0.015} & \textbf{0.915$\pm$ 0.005} \\ 
\bottomrule
\end{tabular}
\label{table_result_2}
\end{adjustbox}
\end{table*}




\subsection{Population Density Prediction}
Table~\ref{table_result_1} presents the performance of population density prediction. 
With three modalities, \modelname $(\mathcal{S} + \mathcal{P} + \mathcal{A})$ and \modelname $(\mathcal{B} + \mathcal{P} + \mathcal{A})$ achieve 1.6\% and 5.9\% improvements in $R^2$ over Urban2Vec and RegionDCL. 
When compared to models using $\mathcal{MB}$ (i.e., MGFN and MVURE), \modelname ($\mathcal{B} + \mathcal{P} + \mathcal{A}$) shows an 8.5\% improvement. \modelname also outperforms the inter-view only models, with a 28.5\% improvement over KnowCL, indicating the importance of incorporating intra-view learning. Compared to GFactorCL, \modelname $(\mathcal{S} + \mathcal{P} + \mathcal{A})$ outperforms by 6.3\% in $R^2$.
With four modalities, \modelname $(\mathcal{S} + \mathcal{B} + \mathcal{P} + \mathcal{A})$ achieves an improvement of $\geq$8.1\% in $R^2$ compared to other baselines that use three modality combinations. 
Notably, \modelname $(\mathcal{S} + \mathcal{B} + \mathcal{P} + \mathcal{A})$ outperforms GFactorCL by 3.4\% in $R^2$. This shows that \modelname captures high-order dependencies without explicitly modeling high-order information, unlike GFactorCL. 
Furthermore, these findings show that the ML-based predictor can filter redundant information in \modelnamex.

\subsection{Crime Rate Prediction} 
Table~\ref{table_result_1} presents the performance of crime rate prediction in NYC. 
\modelname ($\mathcal{B} + \mathcal{P} + \mathcal{A}$) achieves a 5.6\% improvement over RegionDCL, consistently reducing MAE and RMSE scores by at least 5.3 and 4.5, respectively. 
\modelname ($\mathcal{S} + \mathcal{P} + \mathcal{A}$) performs comparably to Urban2Vec in $R^2$ and MAE while lowering RMSE by 9.74. 
Notably, \modelname outperforms GFactorCL by 5.2\% in $R^2$ with $\mathcal{S} + \mathcal{P} + \mathcal{A}$ and 4.8\% with $\mathcal{S} + \mathcal{B} + \mathcal{P} + \mathcal{A}$. Despite not explicitly modeling high-order dependencies like GFactorCL, \modelname achieves superior performance, showing the effectiveness of its pairwise inter-view learning approach.

\subsection{Greenness Score Prediction}
\begin{figure}[h]
    \centering
\includegraphics[width=0.45\textwidth]{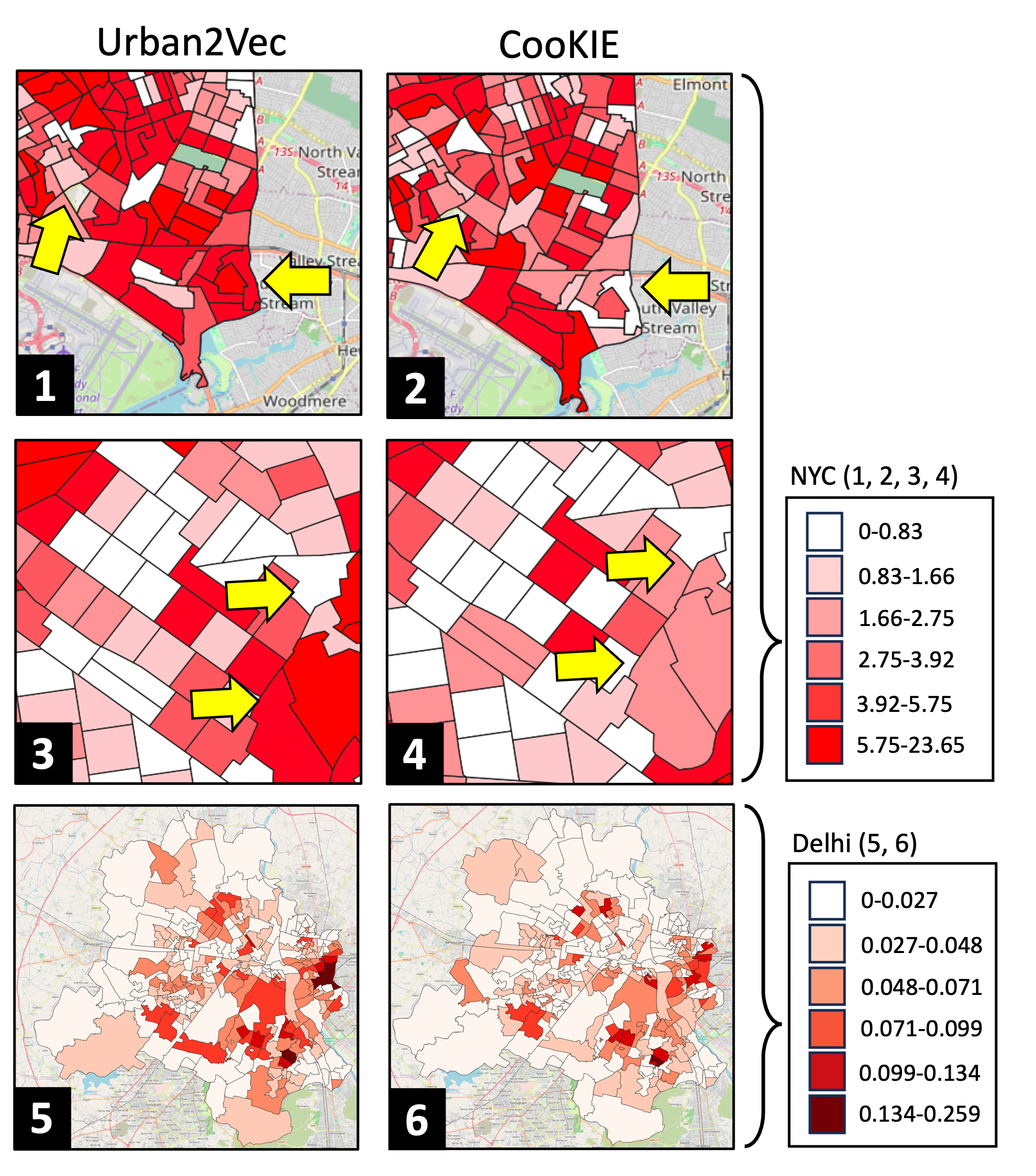}
    \caption{Error distribution in greenness score in NYC and Delhi between Urban2Vec and \modelname ($\mathcal{S}+\mathcal{P}+\mathcal{A}$). The lighter the color (close to white), the smaller the error.}
    \label{fig_result_gvi}
\end{figure}

Table~\ref{table_result_2} presents the performance of greenness score prediction in NYC. 
\modelname $(\mathcal{S} + \mathcal{B} + \mathcal{P} + \mathcal{A})$ outperforms baseline methods, achieving an improvement of $\geq$3.5\% in $R^2$ over RegionDCL, GFactorCL, and MVURE, while showing similar performance to Urban2Vec, with only a 1.2\% difference in $R^2$. 
Figure~\ref{fig_result_gvi} shows the error distribution, where \modelname exhibits smaller errors than Urban2Vec, particularly in non-squared neighborhoods. In comparisons 1 vs. 2 and 3 vs. 4 (highlighted by yellow arrows), Urban2Vec shows higher error rates, likely due to its reliance on spatial proximity for positive sample selection. This approach assumes that nearby regions share similar characteristics, which hold in grid-like urban layouts but fail in irregular neighborhoods. In non-squared areas, curving boundaries can disrupt spatial continuity, making proximity-based representations less reliable. In contrast, \modelname does not rely on spatial proximity for positive samples. Without considering additional geographic information, \modelname achieves lower errors in non-uniform urban environments.

\begin{table}[h]
\centering
\caption{Comparison of \modelname with baseline models for greenness score prediction in Delhi.}
\begin{adjustbox}{width=0.8\columnwidth,center}
  \begin{tabular}{l | c | c} \toprule
  Methods & Modalities & Greenness Score ($R^2$) \\ \midrule
  RegionDCL & $\mathcal{B}$ + $\mathcal{P}$ + $\mathcal{A}$ & 0.273$\pm$0.152\\
  GFactorCL & $\mathcal{B}$ + $\mathcal{P}$ + $\mathcal{A}$ & 0.299$\pm$0.114\\
  \modelname &$\mathcal{B}$ + $\mathcal{P}$ + $\mathcal{A}$& 0.346$\pm$0.069\\
  \midrule
  Urban2Vec & $\mathcal{S}$ + $\mathcal{P}$ + $\mathcal{A}$ & 0.192$\pm$0.051\\
  GFactorCL & $\mathcal{S}$ + $\mathcal{P}$ + $\mathcal{A}$ & 0.290$\pm$0.086\\
  \modelname &$\mathcal{S}$ + $\mathcal{P}$ + $\mathcal{A}$& 0.318$\pm$0.088\\
  \midrule
  \modelname &$\mathcal{S}$ + $\mathcal{B}$ + $\mathcal{P}$ + $\mathcal{A}$ & 0.320$\pm$0.082 \\ \bottomrule
  \end{tabular}
  \label{table_result_delhi}
  \end{adjustbox}
\end{table}
To examine this pattern across different cities, Table~\ref{table_result_delhi} shows $R^2$ for greenness score prediction in Delhi. \modelname ($\mathcal{S}$ + $\mathcal{P}$ + $\mathcal{A}$ and $\mathcal{B}$ + $\mathcal{P}$ + $\mathcal{A}$) significantly outperforms all baselines. The error distribution in Figure~\ref{fig_result_gvi} (5 vs. 6) further shows that \modelname maintains consistently lower prediction errors across the city.


\subsection{Land Use Classification}
Table~\ref{table_result_2} shows that \modelname $(\mathcal{S} + \mathcal{B} + \mathcal{P} + \mathcal{A})$ outperforms all baseline models for land use classification, while \modelname ($\mathcal{S} + \mathcal{P} + \mathcal{A}$) achieves results similar to Urban2Vec. 
This similar performance may stem from how each model incorporates geographic features from $\mathcal{P}$ and $\mathcal{A}$. Urban2Vec calculates spatial proximity as additional geographic information during contrastive learning, whereas \modelname achieves similar results without relying on such explicit spatial relationships.
Compared to GFactorCL ($\mathcal{S} + \mathcal{P} + \mathcal{A}$), \modelname outperforms by 1.7\% in L1 distance and 3.5 in KL-divergence. With $\mathcal{S} + \mathcal{B} + \mathcal{P} + \mathcal{A}$, \modelname achieves further improvements, surpassing GFactorCL by 1.8 in KL-divergence and 0.3\% in cosine similarity. These findings highlight the effectiveness of pairwise inter-view learning in modeling region characteristics without depending on spatial proximity constraints, making \modelname flexible across diverse geographical settings.

\subsection{Case Studies}
We conduct three case studies on NYC data to evaluate \modelnamex's effectiveness. We analyze: (1) model complexity, including parameter efficiency and FLOPs, compared to GFactorCL, (2) the impact of different learning objectives on model performance, and (3) the effect of various modality combinations on model performance.

\subsubsection{Efficiency \& Complexity Analysis}
Table~\ref{table:params} compares \modelname and GFactorCL in terms of parameter count and FLOPs, which directly influence computational cost.
For $\mathcal{B}$ + $\mathcal{P}$ + $\mathcal{A}$ setting, \modelname reduces parameters by 56.7\% and FLOPs by 63.0\%, demonstrating a substantial computational advantage over GFactorCL. 
With $\mathcal{S}$ + $\mathcal{P}$ + $\mathcal{A}$, \modelname reduces parameters by 9.5\% and FLOPs drops to 0.25\%. This is due to the differences in backbone encoders: ResNet50 for $\mathcal{S}$ and an MLP for $\mathcal{B}$. When incorporating all four modalities ($\mathcal{S}$ + $\mathcal{B}$ + $\mathcal{P}$ + $\mathcal{A}$), \modelname learns 10 objectives, whereas GFactorCL learns 15 due to its explicit high-order modeling. As a result, GFactorCL has a 46.3\% increase in parameters, while \modelname improves FLOPs by 1.42\%. This trend shows \modelnamex's efficiency in handling multiple modalities while reducing computational complexity.

\begin{table}[h]
\centering
\caption{Comparison of the number of parameters in \modelname and GFactorCL. \% P-Increase represents the percentage increase in the number of parameters, and \% F-Increase indicates the percentage increase in FLOPs. Both metrics are computed as $(\frac{\text{GFactorCL} - \text{CooKIE}}{\text{CooKIE}}) * 100$.}
\begin{adjustbox}{width=0.99\columnwidth,center}
  \begin{tabular}{l | c | c | c| c| c} \toprule
  Model & Modalities & \# Params & \% P-Increase & \# FLOPs & \% F-Increase \\ \midrule
  \modelname & $\mathcal{B}$ + $\mathcal{P}$ + $\mathcal{A}$ & 13,385,455 & \multirow{2}{*}{56.7\%} & 8.67G & \multirow{2}{*}{63.0\%} \\
  GFactorCL & $\mathcal{B}$ + $\mathcal{P}$ + $\mathcal{A}$ & 20,972,659 & & 14.13 G  \\ \midrule
  \modelname & $\mathcal{S}$ + $\mathcal{P}$ + $\mathcal{A}$ & 96,802,919 & \multirow{2}{*}{9.5\%} & 2,391.23 G & \multirow{2}{*}{0.26\%} \\ 
  GFactorCL & $\mathcal{S}$ + $\mathcal{P}$ + $\mathcal{A}$ & 106,000,347 & & 2,397.45 G \\ \midrule
  \modelname & $\mathcal{S}$ + $\mathcal{B}$ + $\mathcal{P}$ + $\mathcal{A}$ & 107,294,132 & \multirow{2}{*}{46.3\%} & 2,398.53G & \multirow{2}{*}{1.42\%} \\
  GFactorCL & $\mathcal{S}$ + $\mathcal{B}$ + $\mathcal{P}$ + $\mathcal{A}$ & 157,022,271 & & 2,432.59G \\
  \bottomrule
  \end{tabular}
  \label{table:params}
  \end{adjustbox}
\end{table}

\subsubsection{Impact of Intra- and Inter-View Learning on Performance}
We evaluate the contribution of key components in \modelname by removing intra-view learning (\modelnamex-IR) and unique feature extraction in inter-view learning (\modelnamex-UR). Table~\ref{table_result_ablation} reports the $R^2$ results for two downstream tasks: greenness score and crime rate prediction, representing a public health indicator and a commonly used benchmark task. We observe that \modelname consistently outperforms both variants in two tasks.
For greenness score prediction, \modelnamex-UR shows an 8.4\% drop, since greenness is closely tied to $\mathcal{S}$, where visual features are crucial.
For crime rate prediction, \modelnamex-UR reduces performance by 6.9\%, highlighting the importance of unique features in capturing regional characteristics.
Similarly, \modelnamex-IR results in a 2.4\% drop, indicating that intra-view learning captures region-specific representations within each modality. 
These results confirm that both intra-view learning and unique feature extraction in inter-view learning are essential for capturing comprehensive regional representations in multimodal RRL.

\begin{table}[htbp]
\centering
\caption{Test various learning objectives to predict greenness score and crime rate ($R^2$) in NYC. \modelnamex-IR removes intra-view learning from \modelname and \modelnamex-UR removes unique feature learning in inter-view learning.}
\begin{adjustbox}{width=0.85\columnwidth,center}
  \begin{tabular}{l | c | c | c}  \toprule
  Methods & Modalities & Greenness Score & Crime Rate \\ \midrule
  \multirow{2}{*}{\modelnamex-IR} & $\mathcal{S}$ + $\mathcal{P}$ + 
   $\mathcal{A}$ &  0.234$\pm$0.04 & 0.275$\pm$0.03 \\
   & $\mathcal{B}$ + $\mathcal{P}$ + $\mathcal{A}$& 0.259$\pm$0.02 & 0.336$\pm$0.04 \\\midrule
   \multirow{2}{*}{\modelnamex-UR}& $\mathcal{S}$ + $\mathcal{P}$ + 
   $\mathcal{A}$ & 0.166$\pm$0.02 & 0.26$\pm$0.03 \\
   & $\mathcal{B}$ + $\mathcal{P}$ + $\mathcal{A}$& 0.285$\pm$0.02 & 0.32$\pm$0.03 \\ \midrule
  \multirow{2}{*}{\modelname} & $\mathcal{S}$ + $\mathcal{P}$ + $\mathcal{A}$ & 0.250$\pm$0.07 & 0.322$\pm$0.03 \\
  & $\mathcal{B}$ + $\mathcal{P}$ + $\mathcal{A}$ & 0.292$\pm$0.01 & 0.346$\pm$0.02 \\
   \bottomrule
  \end{tabular}
  \label{table_result_ablation}
\end{adjustbox}
\end{table}

\subsubsection{Impact of Modality Combinations on Performance}
\begin{figure}[h]
    \centering
    \includegraphics[width=0.5\textwidth]{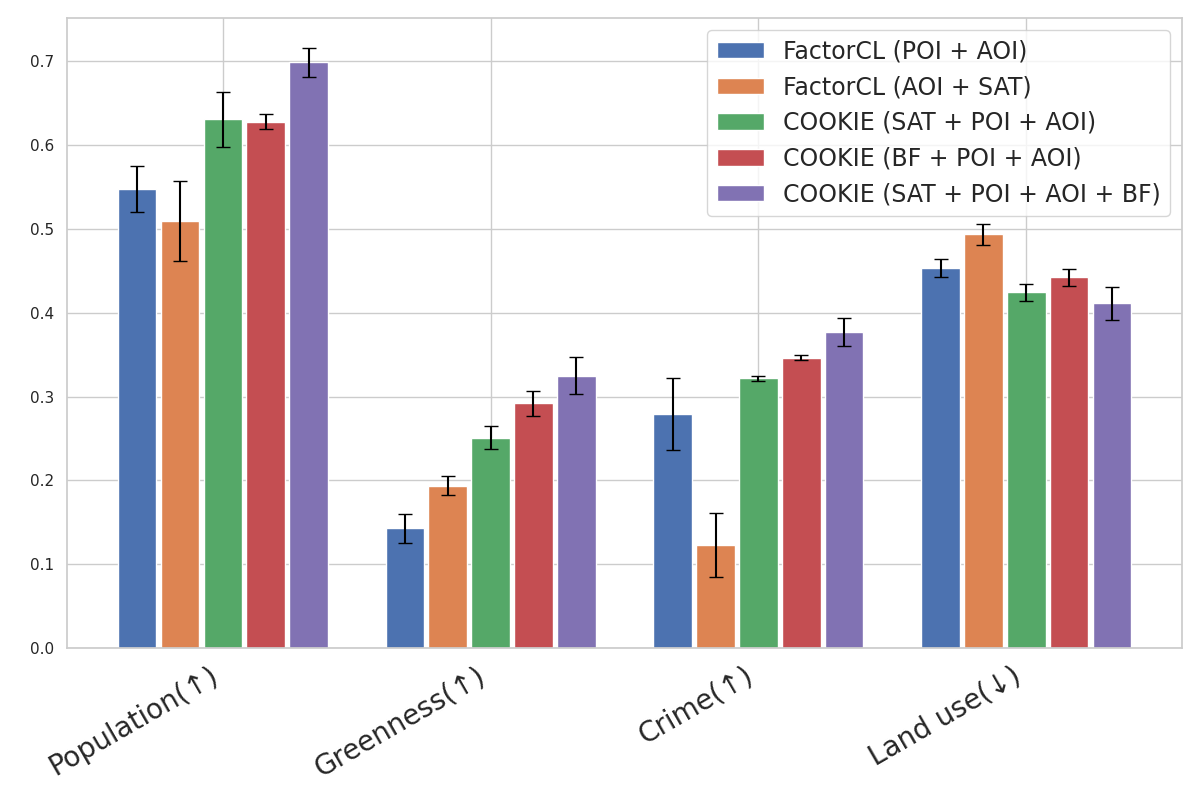}
    \vspace{0.5em}
    \caption{Effect of Modalities. We report $R^2$ for population density, greenness score, and crime rate, and $L1$ on land use classification to show the effect of modalities.}
    \label{all_2}
\end{figure}

Figure~\ref{all_2} shows how different modality combinations impact the performance of FactorCL and \modelnamex. 
The results reveal a consistent pattern: adding $\mathcal{S}$ (satellite imagery, green) and $\mathcal{B}$ (building footprints, red) significantly improves performance across most tasks. 
\modelname achieves the highest scores when incorporating all four modalities ($\mathcal{S} + \mathcal{B}+ \mathcal{P}+ \mathcal{A}$), consistently outperforming combinations that use only three modalities across all tasks. 
In contrast, FactorCL struggles when limited to two modalities, showing weaker performance than \modelnamex. This indicates that FactorCL has difficulty capturing comprehensive region representation without diverse geospatial modalities. This performance gap highlights the importance of using multiple modalities, as fewer datasets restrict the model’s downstream task performance in RRL.

\section{Related Work}
RRL aims to map raw geographic data to feature vectors that capture spatial information and regional attributes. 
Early RRL approaches~\cite{wang2017region} use a single data modality, specifically human mobility data, to learn region embeddings but could benefit from incorporating additional modalities.
Subsequent methods incorporate multiple data sources, including mobility, point-of-interest (POI), and satellite imagery, to enrich region representations. CGAL~\cite{zhang2019unifying} constructs region graphs from taxi trajectories and POI data, RegionEncoder~\cite{jenkins2019unsupervised} combines satellite imagery, POI data, and mobility flows through a graph-based approach, and MV-PN~\cite{fu2019efficient} uses mobility data and geographic distance to build intra-region POI networks. MVURE~\cite{mvure} uses a graph-attention network to model region relationships from POI and mobility data, while more recent attention-based methods like MGFN~\cite{wu2022multi_mgfn}, HAFusion~\cite{sun2024urban}, and HREP~\cite{zhou2023heterogeneous} highlight important features in each modality. These approaches rely on concatenation, weighted summation, or view-wise attention, which can overlook inter-modal dependencies when not explicitly designed to capture them.

Contrastive learning (CL) provides a powerful strategy for aligning representations across data modalities by forming positive and negative sample pairs~\cite{contrastivesurvey}. SimCLR~\cite{simclr} and CLIP~\cite{clip} illustrate how such alignment works for image and text data, and geospatial methods like KnowCL~\cite{liu2023knowledge}, RegionDCL~\cite{li2023urban}, ReMVC~\cite{tkde}, and ReCP~\cite{li2024urban} uses CL to integrate POI data, mobility, and satellite imagery. However, these methods often focus on shared information when aligning modality in inter-view learning. FactorCL~\cite{factorcl} addresses this issue by modeling both shared and unique information between two modalities, but its complexity escalates with additional modalities. In contrast, \modelname uses pairwise inter-view learning to capture both shared and unique information across multiple modalities while reducing the number of learning objectives, offering a more scalable solution for multimodal RRL.

\section{Conclusion, Limitations, and Future work} 
We propose \modelnamex, a novel multimodal region representation learning method that applies information factorization to more than two geospatial data. Based on the factorization approach, \modelname can capture modality-specific and shared information during data alignment in contrastive learning.
Further, \modelname uses pairwise inter-view learning to model shared information efficiently, avoiding explicit high-order dependency modeling. This method enables \modelname to achieve state-of-the-art performance across multiple downstream tasks in NYC, United States, and Delhi, India. 
Although our pairwise learning approach is effective for the chosen modalities, its effectiveness for other geospatial data types remains uncertain.
Future work will evaluate its applicability to diverse datasets, such as road networks and climate data.








\bibliographystyle{ACM-Reference-Format}
\bibliography{ref}


\end{document}